\documentclass{article}
\usepackage{ijcai26}

\usepackage{times}
\usepackage{soul}
\usepackage{url}
\usepackage[hidelinks]{hyperref}
\usepackage[utf8]{inputenc}
\usepackage[small]{caption}
\usepackage{graphicx}
\usepackage{amsmath}
\usepackage{amsthm}
\usepackage{booktabs}
\usepackage{algorithm}
\usepackage{algorithmic}
\usepackage[switch]{lineno}

\newtheorem{theorem}{Theorem}

\usepackage{multirow}
\usepackage{adjustbox}
\usepackage{caption}
\usepackage{makecell}
\usepackage{graphicx}
\usepackage{amsmath, amssymb, amsthm}

\title{BetaEdit: Null-Space Constrained Sequential Model Editing}

\author{
Bingqing Liu
\and
Wei Liu\thanks{Corresponding author. Accepted by IJCAI 2026.}\and
Yuhua Li\\
\affiliations
Huazhong University of Science and
Technology, Wuhan, China\\
\emails
\{liubingqing24, idc\_lw, idcliyuhua\}@hust.edu.cn
}

\begin{document}

\maketitle

\begin{abstract}
Null-space-based methods have garnered considerable attention in model editing by constraining updates to the null space of the pre-existing knowledge representation, thereby preserving the model's original behavior. However, in practice these methods rely on an approximate null space—leading to \textit{knowledge leakage}—and further suffer from severe performance degradation during sequential editing. Recent work shows that \textit{history-aware} editing strategies can empirically mitigate this decline, yet the underlying reason remains unclear. In this paper, we first expose the knowledge leakage inherent in existing null-space approaches and then analyze why history-aware updates effectively preserve both editing  performance and general capabilities during long-horizon editing. Building on these insights, we propose \textsc{BetaEdit}, a refined framework that effectively controls the  knowledge leakage and integrates history-aware updates into the null-space paradigm. Extensive experiments on three large language models across two standard benchmarks show that \textsc{BetaEdit} consistently outperforms prior methods in the challenging regime of massive-scale sequential editing. Code is available at: \url{https://github.com/lbq8942/BetaEdit}.
\end{abstract}

\section{Introduction}
\label{sec:intro}

Model editing seeks to locally modify a pre-trained model's behavior—e.g., updating a factual belief or correcting a misconception—without retraining from scratch or compromising its general capabilities \cite{jiang2025neuron,liumitigating,xie2025revealing,yang2025mirage,liuunlocking}. A central challenge in this task is balancing two competing objectives: (1) faithfully incorporating new knowledge, and (2) preserving the vast pre-trained knowledge that underpins the model’s general capability. Early approaches often rely on explicit regularization to protect pre-existing knowledge \cite{meng2022locating,mengmass,gupta2024unified}, but such methods struggle to simultaneously achieve high edit accuracy and strong edit locality \cite{fangalphaedit}.

\begin{figure}[htbp]
  \centering
  \includegraphics[width=0.95\linewidth]{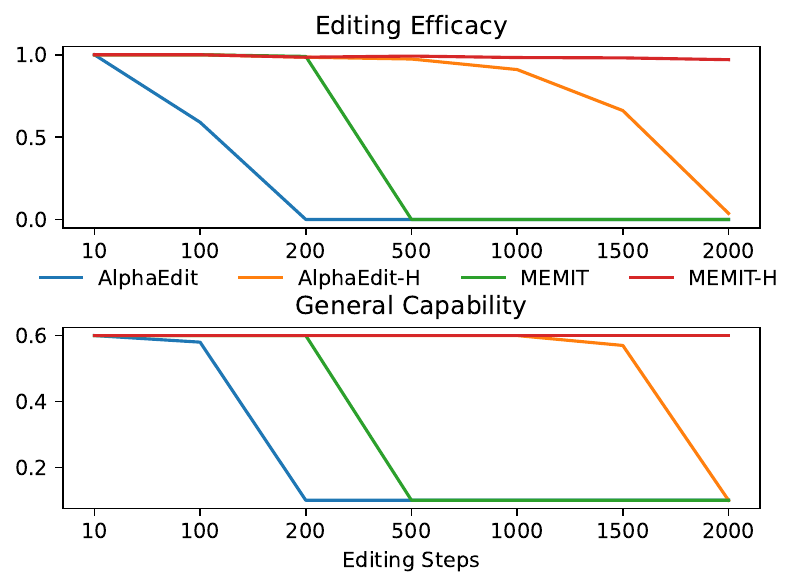}
\caption{Editing efficacy and general capability (measured by Massive Multitask Language Understanding, MMLU) of various editing methods over 2,000 sequential edits on \textsc{LLaMA}3-8B using the CounterFact dataset. Markers with a ``H'' suffix indicate history-aware editing, while those without denote history-agnostic editing.}
  \label{fig:hist_aware}
\end{figure}

To circumvent this trade-off, null-space-based editing methods~\cite{fangalphaedit} propose a theoretically elegant solution: perform edits exclusively within the null space defined by pre-existing knowledge space. By construction, such edits are intended to induce zero change on the model’s outputs for unedited inputs, thereby preserving pre-trained knowledge perfectly without explicit regularization. However, we observe that this ideal behavior exists only in theory and fails in practice. We identify the root cause: the “null space” used in practice is not exact, but an \textit{approximate} (or \textit{pseudo}) null space obtained via truncated singular value decomposition. This approximation inevitably introduces \textit{knowledge leakage}, i.e., small but non-zero perturbations to pre-trained knowledge, which accumulates over sequential edits and eventually corrupts both editing fidelity and the model’s general capabilities. 

Recent history-aware editing strategy~\cite{fangalphaedit,fanghippocampal}, which explicitly regularizes new updates based on past edits, has been shown to significantly mitigate this degradation. As shown in Figure~\ref{fig:hist_aware}, history-aware regularization substantially improves edit efficacy and preserves the model's general capability during sequential editing. The gain in edit efficacy is expected due to the protection of past edits; the concurrent preservation of the model's general capability, however, is counterintuitive. How the history-aware regularization---a single mechanism---achieves both effects during sequential editing remains unclear.

In this paper, we bridge these gaps. First, we formally identify and quantify the knowledge leakage inherent in current null-space methods (Section~\ref{sec:leakage}). Second, we provide a principled theoretical analysis (Section~\ref{sec:analysis}) showing that history-aware updates implicitly regularize the edit trajectory by reducing interference among sequential edits and constraining overall weight perturbation, thereby better preserving both editing fidelity and general capabilities. Building on these insights, we propose \textsc{BetaEdit} (Section~\ref{sec:method}), a refined framework that (1) mitigates knowledge leakage by reintroducing a penalty term  specifically designed to compensate for the imperfection of the pseudo-null space, and (2) seamlessly integrates history-aware updates into the null-space paradigm.  Experiments on three large language models and two standard editing benchmarks demonstrate that \textsc{BetaEdit} remains effective even after 10,000 sequential edits, whereas existing editors typically fail, severely corrupting the model’s general capabilities and achieving near-zero editing efficacy.

\section{Related Work}

Model editing techniques are broadly classified according to whether they modify the original parameters of the pre-trained model.

\textbf{Parameter-preserving editing} methods aim to update or correct model behavior without altering the base weights. These approaches generally fall into two categories. The first introduces auxiliary modules to inject or override knowledge dynamically. For example, SERAC~\cite{mitchell2022memory} employs an external memory store together with a lightweight corrective model; GRACE~\cite{hartvigsen2023aging} stores knowledge in discrete codebooks; while methods like CaliNet~\cite{dong2022calibrating}, T-Patcher~\cite{huangtransformer}, MELO~\cite{yu2024melo}, and WISE~\cite{wang2024wise} incorporate trainable modules (e.g., LoRA adapters, or side memory) to store new knowledge. The second category relies on contextual prompting strategies—such as IKE~\cite{zheng2023can}, Deck~\cite{bi2025decoding}, and DecKER~\cite{wang2025decoupling}—which steer the model toward desired outputs at inference time through carefully designed prompts.

In contrast, \textbf{parameter-modifying editing} directly updates the model’s internal weights. These methods typically follow one of two paradigms. The first leverages meta-learning frameworks, where a hypernetwork is trained to predict localized parameter adjustments. Notable examples include MEND~\cite{mitchellfast}, MALMEN~\cite{tanmassive}, InstructEdit~\cite{zhang2024instructedit}, and RLEdit~\cite{lireinforced}. The second adopts a ``locate-then-edit'' strategy: it first pinpoints the specific activations or parameter subsets responsible for a target fact—often using gradient-based attribution or causal tracing \cite{zhou2025editing}—and then applies precise modifications. Representative works include ROME~\cite{meng2022locating}, MEMIT~\cite{mengmass}, EMMET~\cite{gupta2024unified}, and AlphaEdit~\cite{fangalphaedit}. Recent efforts within this paradigm also focus on mitigating model degradation during sequential editing. Techniques along this line include post-editing weight regularization~\cite{gu2024model,maperturbation,li2025adaedit,liuedit,qiao2025wasserstein} and history-aware regularization strategies that explicitly account for past edits~\cite{fangalphaedit,fanghippocampal,guotowards}.

\section{Knowledge Leakage in Null-Space Constrained Editing}
\label{sec:leakage}

We consider the \textit{locate-then-edit} paradigm for model editing \cite{meng2022locating}, which proceeds in two steps: (1) identifying a critical layer that stores factual knowledge—commonly the second linear layer in the MLP of early transformer blocks—and (2) updating its weight matrix $\mathbf{W}$ to incorporate new information while preserving pre-existing knowledge. This is formalized as a regularized least-squares problem~\cite{mengmass}:
\begin{equation}
\begin{split}
\Delta = \arg\min_{\tilde{\Delta}} & \left\|(\mathbf{W} + \tilde{\Delta})\mathbf{K}_1 - \mathbf{V}_1\right\|_F^2 \\
&+ \lambda_1\left\|(\mathbf{W} + \tilde{\Delta})\mathbf{K}_0 - \mathbf{V}_0\right\|_F^2,
\label{eq:original_lte}
\end{split}
\end{equation}
where $(\mathbf{K}_0, \mathbf{V}_0)$ denotes pre-trained knowledge (satisfying $\mathbf{W}\mathbf{K}_0 = \mathbf{V}_0$ by construction), and $(\mathbf{K}_1, \mathbf{V}_1)$ represents the new factual update to be injected. The regularization parameter $\lambda_1 > 0$ controls the trade-off between edit accuracy and preservation of pre-trained knowledge. The closed-form solution to Eq.~\eqref{eq:original_lte} is given by:
\begin{equation}
\Delta = \mathbf{R}_1 \mathbf{K}_1^\top \left( \lambda_1 \mathbf{K}_0 \mathbf{K}_0^\top + \mathbf{K}_1 \mathbf{K}_1^\top \right)^{-1},
\label{eq:memit_closed}
\end{equation}
with $\mathbf{R}_1 = \mathbf{V}_1 - \mathbf{W}\mathbf{K}_1$ denoting the residual of the new knowledge.

To avoid compromising pre-trained knowledge, AlphaEdit~\cite{fangalphaedit} reparameterizes the update as $\tilde{\Delta} \mathbf{P}$, where $\mathbf{P}$ is a projection matrix onto the null space of $\mathbf{K}_0$, i.e., $\mathbf{P} \mathbf{K}_0 = \mathbf{0}$. With this constraint, the second term in Eq.~\eqref{eq:original_lte} vanishes identically, yielding the simplified objective:
\begin{equation}
\Delta = \arg\min_{\tilde{\Delta}\mathbf{P}} \left\|(\mathbf{W} + \tilde{\Delta}\mathbf{P})\mathbf{K}_1 - \mathbf{V}_1\right\|_F^2.
\label{eq:alphaedit_obj}
\end{equation}
This yields the following closed-form update:
\begin{equation}
\Delta = \mathbf{R}_1 \mathbf{K}_1^\top \mathbf{P} \left( \lambda_2 \mathbf{I} + \mathbf{K}_1 \mathbf{K}_1^\top \mathbf{P} \right)^{-1},
\label{eq:alphaedit_closed}
\end{equation}
where $\lambda_2 > 0$ ensures numerical invertibility.

\begin{figure}[htbp]
  \centering
  \includegraphics[width=0.95\linewidth]{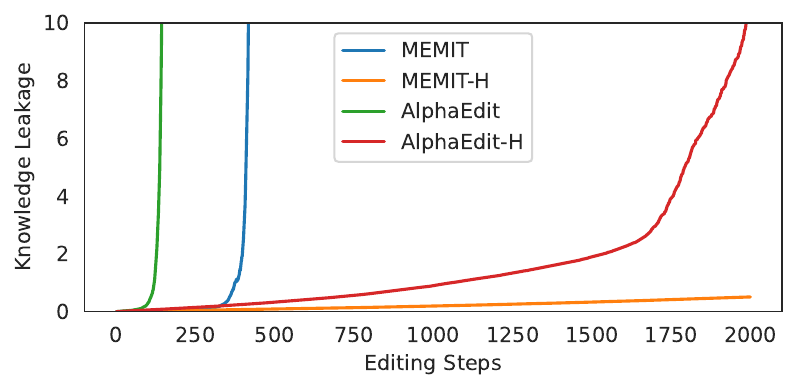}
  \caption{Knowledge leakage of existing editing methods, evaluated on \textsc{LLaMA}3-8B and CounterFact.}
  \label{fig:leakage}
\end{figure}
In theory, the null-space formulation guarantees perfect preservation of pre-trained knowledge: for any sequence of edits $\{\Delta^{(t)}\}_{t=1}^T$, the cumulative update satisfies $(\mathbf{W} + \sum_{t=1}^T \Delta^{(t)}) \mathbf{K}_0 = \mathbf{V}_0$, since each $\Delta^{(t)} \mathbf{K}_0 = \mathbf{0}$. In practice, however, we observe a violation of this equality. To quantify the deviation, we define \textit{knowledge leakage} as:
\[
\mathrm{KL} = \left\| (\mathbf{W} + \Delta) \mathbf{K}_0 - \mathbf{V}_0 \right\|_F,
\]
which measures unintended perturbations to pre-trained knowledge after edit. Figure~\ref{fig:leakage} shows the $\mathrm{KL}$ with respect to the cumulative update $\sum_{t=1}^T \Delta^{(t)}$ over $T=2000$ sequential edits. Contrary to theoretical expectations, both null-space methods, AlphaEdit and AlphaEdit-H, exhibit non-zero and growing knowledge leakage, often exceeding that of baseline methods such as MEMIT \cite{mengmass}. This directly contradicts the core promise of null-space editing and explains the catastrophic model corruption observed in long-horizon settings (cf.~Figure~\ref{fig:hist_aware}).

The root cause lies in the \textit{approximate} nature of the employed null space. 
In practice, $\mathbf{K}_0 \mathbf{K}_0^\top$ is typically full-rank~\cite{gupta2025efficient}, which implies that all singular values of $\mathbf{K}_0 \mathbf{K}_0^\top$ are positive, 
rendering the true null space trivial (i.e., $\{\mathbf{0}\}$). 
To address this, prior work treat singular values below a threshold (e.g., $0.02$)  as 0 \cite{fangalphaedit}.
This approximation implies that $\mathbf{P} \mathbf{K}_0 \neq \mathbf{0}$, causing each edit to introduce a small leakage that accumulates over time.

\section{Theoretical Analysis: History-Aware Updates Reduce Weight Perturbation}
\label{sec:analysis}

Naively applying the update rules in Eq.~\ref{eq:memit_closed} or Eq.~\ref{eq:alphaedit_closed} during sequential editing often leads to catastrophic degradation—both in editing efficacy and in the model’s general capabilities \cite{gu2024model}. Recent work by \cite{fangalphaedit,fanghippocampal} empirically demonstrates that this degradation can be substantially mitigated by explicitly preserving \textit{previously edited facts} when performing new edits.

Concretely, under a history-aware editing mechanism, the MEMIT-style weight update at step \( t \) takes the form:
\begin{equation}
    \Delta_t = r_t k_t^\top \left(k_t k_t^\top + \sum_{s=1}^{t-1} k_s k_s^\top + \lambda_1 \mathbf{K}_0 \mathbf{K}_0^\top \right)^{-1},
    \label{eq:history_aware_update}
\end{equation}
where the key vectors from previous edits, \(k_1, \dots, k_{t-1}\), are explicitly incorporated into the \(t\)-th update. In contrast, the vanilla (history-agnostic) approach uses only:
\begin{equation}
    \Delta_t = r_t k_t^\top \left(k_t k_t^\top + \lambda_1 \mathbf{K}_0 \mathbf{K}_0^\top \right)^{-1}.
    \label{eq:history_agnostic_update}
\end{equation}
Compared to Eq.~\eqref{eq:memit_closed}, we use lowercase \(r_t\) and \(k_t\) here to denote vectors (rather than matrices), and the subscript \(t\) emphasizes the sequential editing setting.

While the improvement in editing accuracy is intuitive—since past edits are directly protected—it is less obvious why general capabilities are simultaneously preserved. In what follows, we provide a theoretical analysis showing that the history-aware mechanism not only safeguards previously edited knowledge but also better preserves the model’s general capabilities during sequential editing, in contrast to the vanilla (history-agnostic) approach.

To ensure a clean and meaningful comparison, we first exclude certain trivial cases. For instance, if the residual vector $r_t$ is zero, no actual edit occurs; if two edits introduce conflicting updates (e.g., “the capital of France is Paris” vs. “the capital of France is London”), they may cancel each other out, potentially restoring the model to its initial state without degrading general capabilities. Such scenarios do not reflect the steady growth of knowledge leakage observed in Figure \ref{fig:leakage}. Therefore, we assume that the sequence of weight perturbations $\{\Delta_t\}$ is \emph{non-conflicting}, i.e., $\langle \Delta_i, \Delta_j \rangle_F \geq 0$ for all $i, j$, which ensures  non-decreasing knowledge leakage over time (see the supplementary material for details).

Under this editing scenario, the following theorem formalizes the advantage of the history-aware approach:

\begin{theorem}[Reduced Weight Perturbation]
\label{thm:single_step_suppression}
Let \(\Delta_t^{(A)}\) and \(\Delta_t^{(B)}\) denote the weight perturbations at the \(t\)-th editing step for the vanilla (history-agnostic) method and the history-aware method, respectively. 
For any \(T > 1\), the cumulative weight perturbation of the history-aware method is smaller in Frobenius norm than that of the vanilla method:
\[
\left\| \sum_{t=1}^T \Delta_t^{(B)} \right\|_F <  \left\| \sum_{t=1}^T \Delta_t^{(A)} \right\|_F.
\]
\end{theorem}

\paragraph{Proof sketch.}
The key insight is that the history-aware update $\Delta_t^{(B)}$ explicitly incorporates previously edited directions $\{k_s\}_{s < t}$ via a Gram matrix that includes $\sum_{s=1}^{t-1} k_s k_s^\top$. This effectively projects the new update onto a subspace orthogonal to past edit directions, thereby reducing interference with earlier updates, i.e., $\langle \Delta_t^{(B)}, \Delta_s^{(B)} \rangle_F \le \langle \Delta_t^{(A)}, \Delta_s^{(A)} \rangle_F$  for all  $ s<t $, and yielding a smaller cumulative weight perturbation. Full details are provided in the supplementary material. 

\begin{figure*}[htbp]
  \centering
  \includegraphics[width=0.95\linewidth]{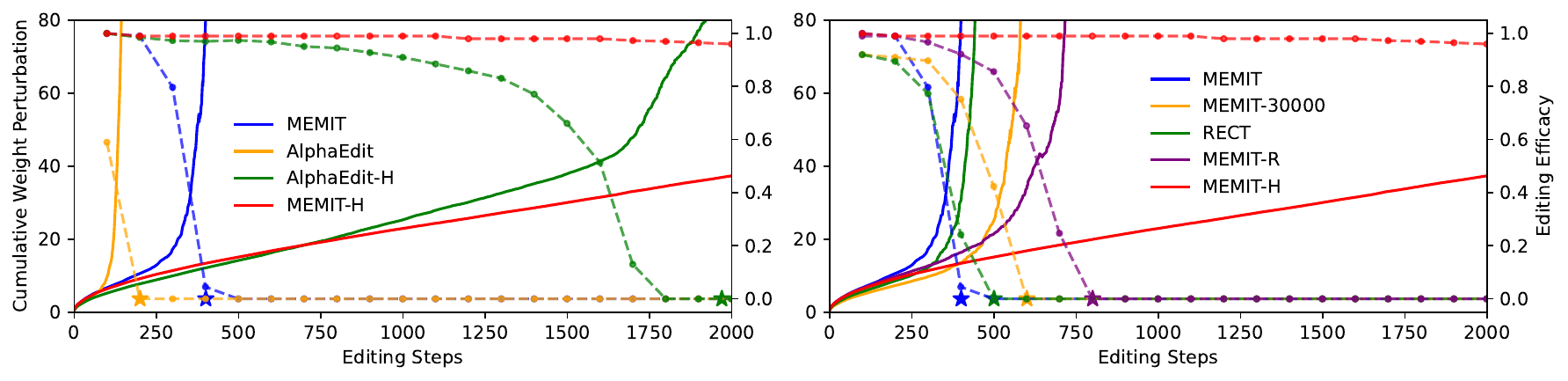}
\caption{Comparison of cumulative weight perturbation between history-agnostic and history-aware editing, evaluated on \textsc{LLaMA}3-8B and the CounterFact dataset. The solid line shows the evolution of the Frobenius norm of the cumulative weight perturbation over 2000 editing steps (left y-axis), while the corresponding editing efficacy is indicated by the dashed line (right y-axis). Large cumulative weight perturbations degrade the model's general capabilities (measured by MMLU), with stars marking the onset of complete corruption.}
  \label{fig:cumulative_norm}
\end{figure*}

Figure~\ref{fig:cumulative_norm}(a) illustrates the evolution over time of the Frobenius norm of the cumulative weight update at each editing step. The history-aware method consistently suppresses the magnitude of perturbation more effectively than the history-agnostic method. This suppression becomes increasingly pronounced as the number of edits grows, since the likelihood of overlap between new and past edits increases. The history-aware method is able to orthogonalize these overlapping components, whereas the history-agnostic method conducts independent updates, leading to an accumulation of interference and an eventual explosion in cumulative weight perturbation, which corrupts the model's general  capabilities (cf.~Figure~\ref{fig:hist_aware}).

To further demonstrate the effectiveness of history-aware editing in suppressing weight perturbation, we evaluate several history-agnostic variants that attempt to reduce per-step weight perturbation:

\begin{itemize}
  \item \textbf{MEMIT-30000}: Increases the regularization strength \(\lambda_1\) from 15,000 to 30,000, thereby shrinking the magnitude of each update \(\Delta_t\).
  \item \textbf{MEMIT-R}: Replaces the historical keys \(\{k_1, \dots, k_{t-1}\}\) in Eq. (\ref{eq:history_aware_update}) with random keys \(\{k_1^{r}, \dots, k_{t-1}^{r}\}\), which strengthens the regularization but lacks meaningful alignment with historical edits.\footnote{Since historical keys are derived from the subject token of each historical edit \cite{fangalphaedit}, the random keys are sampled from subject tokens of randomly selected edits.}
  \item \textbf{RECT} \cite{gu2024model}: Applies a sparsification strategy that zeroes out small-magnitude entries in the update matrix \(\Delta_t\), retaining only the largest components.
\end{itemize}

The results, shown in Figure~\ref{fig:cumulative_norm}(b), indicate that while these strategies somewhat alleviate cumulative weight perturbation, they incur a notable trade-off in terms of editing performance.





\section{Method}
\label{sec:method}

As revealed in Section~\ref{sec:leakage}, existing null-space-based editing methods suffer from non-negligible \textit{knowledge leakage}, which leads to cumulative corruption of the pretrained knowledge over sequential edits. To address this, we propose to explicitly account for the leakage during weight update. Specifically, instead of discarding the preservation term as in Eq.~\eqref{eq:alphaedit_obj}, we reintroduce it into the objective optimization:
\begin{equation}
\begin{split}
\Delta = \arg\min_{\tilde{\Delta}\mathbf{P}} \, & \left\|(\mathbf{W} + \tilde{\Delta}\mathbf{P})\mathbf{K}_1 - \mathbf{V}_1\right\|_F^2 \\
 + \lambda_1 &\left\| (\mathbf{W} + \tilde{\Delta}\mathbf{P}) \mathbf{K}_0 - \mathbf{V}_0 \right\|_F^2,
\end{split}
\label{eq:our_obj_single}
\end{equation}
which yields the following closed-form solution:
\begin{equation}
\Delta = \mathbf{R}_1 \mathbf{K}_1^\top \mathbf{P} \left( \lambda_2 \mathbf{I} + (\mathbf{K}_1 \mathbf{K}_1^\top + \lambda_1 \mathbf{K}_0 \mathbf{K}_0^\top) \mathbf{P} \right)^{-1},
\label{eq:our_closed_single}
\end{equation}
where $\lambda_2 > 0$ ensures numerical stability (set to 10 following \cite{fangalphaedit}), and $\lambda_1 > 0$ balances edit accuracy against knowledge leakage control.

Extending this to the sequential editing setting, we incorporate both pre-trained knowledge and all previously edited facts into the protection set. The update at step $t$ is given by:
\begin{equation}
\Delta_t = \mathbf{R}_t \mathbf{K}_t^\top \mathbf{P}_t \left( \lambda_2 \mathbf{I} + (\mathbf{K}_t \mathbf{K}_t^\top + \mathbf{C}^{(t)}) \mathbf{P}_t \right)^{-1},
\label{eq:our_closed_seq}
\end{equation}
where $\mathbf{R}_t = \mathbf{V}_t - \mathbf{W} \mathbf{K}_t$ and $
\mathbf{C}^{(t)} = \lambda_1 \mathbf{K}_0 \mathbf{K}_0^\top + \sum_{i=1}^{t-1} \mathbf{K}_i \mathbf{K}_i^\top $
aggregates the representations of all protected knowledge up to step $t$, yielding a \textit{history-aware} update.

Motivated by the theoretical analysis in Section~\ref{sec:analysis}—which shows that preserving editing history protects both editing performance and the model's general capabilities—we extend the null-space protection beyond the original knowledge $\mathbf{K}_0$ to also cover all previously edited knowledge $\{\mathbf{K}_i\}_{i=1}^{t-1}$. That is, instead of constructing the projector based solely on $\mathbf{K}_0 \mathbf{K}_0^\top$ as in prior work \cite{fangalphaedit}, we define $\mathbf{P}_t$ as the orthogonal projector onto the  null space of $\mathbf{C}^{(t)}$.

Concretely, we compute the eigendecomposition of $\mathbf{C}^{(t)}=\mathbf{U} \boldsymbol{\Lambda} \mathbf{U}^\top$ and select the columns of $\mathbf{U}$ corresponding to eigenvalues smaller than a threshold $\epsilon$ (e.g., $\epsilon = 0.02$). Denoting this subset of eigenvectors as $\mathbf{U}_{\text{small}}$, we form the projection matrix as $\mathbf{P}_t =\mathbf{U}_{\text{small}} \mathbf{U}_{\text{small}}^\top.
$
By construction, $\mathbf{P}_t$ suppresses directions in which the model has already encoded factual knowledge, ensuring that new updates minimally perturb both pre-trained and edited facts.
However, applying the null-space constraint to the editing history necessitates recomputing $\mathbf{P}_t$ at each edit step, which incurs substantial computational overhead. To balance efficiency and protection, we adopt a periodic refresh strategy: $\mathbf{P}_t$ is updated only every $\tau$ edits (e.g., $\tau = 1,000$). In the limiting case where $\tau \to \infty$, our method reduces to standard null-space editing that protects only the pretrained knowledge.

\section{Experiments}
\subsection{Experimental Setup}
\label{sec:setup}
\begin{table*}[h]
\centering
\scriptsize
\setlength{\tabcolsep}{4pt}
\begin{tabular}{l l c c c c c cc c c c c cc c c c c c}
\toprule
\multirow{3}{*}{\textbf{Method}} & 
& \multicolumn{9}{c}{\textbf{CounterFact}} 
& \multicolumn{9}{c}{\textbf{ZsRE}} \\
\cmidrule(lr){3-11} \cmidrule(lr){12-20}
&&\multicolumn{3}{c}{\textbf{$T$ = 300}}&\multicolumn{3}{c}{\textbf{$T$ = 5000}}&\multicolumn{3}{c}{\textbf{$T$ = 10000}}&\multicolumn{3}{c}{\textbf{$T$ = 300}}&\multicolumn{3}{c}{\textbf{$T$ = 5000}}&\multicolumn{3}{c}{\textbf{$T$ = 10000}}\\
\cmidrule(lr){3-5} \cmidrule(lr){6-8}  \cmidrule(lr){9-11} \cmidrule(lr){12-14} \cmidrule(lr){15-17} \cmidrule(lr){18-20} 
& 
& \textbf{Eff.}& \textbf{Gen.} & \textbf{Spe.}
& \textbf{Eff.} & \textbf{Gen.} & \textbf{Spe.}& \textbf{Eff.} & \textbf{Gen.} & \textbf{Spe.} & \textbf{Eff.} & \textbf{Gen.} & \textbf{Spe.} & \textbf{Eff.} & \textbf{Gen.} & \textbf{Spe.} & \textbf{Eff.} & \textbf{Gen.} & \textbf{Spe.}  \\
\midrule
Pre-edited & &0.30&0.20&15.0& 0.30&0.50 &13.9& 0.30 &0.40&14.0  &  31.6& 32.3&34.8& 32.3 &32.6&35.1  &32.1  &32.2 &34.8 \\
\midrule
FT& \multirow{10}{*}{\rotatebox{90}{Qwen3-4B-Instruct}} 
& 43.6 & 4.10 & 3.20 & 0.00 & 0.00 & 0.00& 0.00 & 0.00 & 0.00
& 38.5 & 7.20 &11.3 & 0.00 & 0.00 & 0.00& 0.00 & 0.00 & 0.00\\
RLEdit&  
& 76.5 & 34.4 & 7.00& 0.00 & 0.00 & 0.00& 0.00 & 0.00 & 0.00 
& 78.8 & 55.7 & 24.3 & 0.00 & 0.00 & 0.00& 0.00 & 0.00 & 0.00  \\
RECT&   
& 80.4 & 53.8 & 8.70 & 0.00 & 0.00 & 0.00& 0.00 & 0.00 & 0.00  
& 83.4 & 70.6 & 28.7& 0.00 & 0.00 & 0.00& 0.00 & 0.00 & 0.00 \\
PRUNE&  
& 85.0 & 55.5 & 8.30& 0.00 & 0.00 & 0.00& 0.00 & 0.00 & 0.00
& 89.3 & 76.0 & 32.0& 0.00 & 0.00 & 0.00& 0.00 & 0.00 & 0.00  \\
AdaEdit&  
& 88.4 & 59.5 & 8.10& 0.00 & 0.00 & 0.00& 0.00 & 0.00 & 0.00 
& 85.0 & 73.0 & 28.4& 0.00 & 0.00 & 0.00& 0.00 & 0.00 & 0.00 \\
\cmidrule(lr){3-20} 
SimIE&  
& 95.0 & 71.3 & 11.8 & 0.00 & 0.00 & 0.00 & 0.00 & 0.00 & 0.00 
& 95.2 & 90.6 & 34.3& 0.00 & 0.00 & 0.00& 0.00 & 0.00 & 0.00  \\
EMMET&  
& 98.5 & \underline{74.0} & 11.7 & 0.00 & 0.00 & 0.00 & 0.00 & 0.00 & 0.00 
& \underline{98.5} & 93.2 & 36.8 & 0.00 & 0.00 & 0.00 & 0.00 & 0.00 & 0.00 \\
PMET&   
& \textbf{98.8} & 72.4 & 12.6 & 82.0 & 55.2 & 7.20 & 0.00 & 0.00 & 0.00
& \textbf{98.6} & \textbf{94.0} & \textbf{37.3} & 88.2 & 80.3 &35.5 & 0.00 & 0.00 & 0.00 \\
MEMIT&  
& 97.0 & 70.8 & \textbf{13.2} & \underline{85.3} & \underline{58.3} & \textbf{8.10} & 0.00 & 0.00 & 0.00
& 96.5 & 93.0 & \underline{37.1} & \underline{95.5} & \underline{89.9} & \textbf{38.9} & \underline{90.0} & \underline{81.2} & \underline{36.3} \\
AlphaEdit& 
& \underline{98.5} & \textbf{74.1} & 12.8 & 0.00 & 0.00 & 0.00 & 0.00 & 0.00 & 0.00 
& 97.9 & \underline{93.3} & 37.0 & 89.1 & 84.0 & 37.4 & 0.00 & 0.00 &0.00 \\
BetaEdit& 
& 98.0 & 73.5 & \underline{13.0} & \textbf{90.3} & \textbf{60.1} & \underline{8.00} & \textbf{78.5} & \textbf{54.8} & \textbf{3.80} 
& 97.5 & 93.0 & 36.7 & \textbf{96.7} & \textbf{90.9} & \underline{38.1} & \textbf{92.1} & \textbf{84.0} & \textbf{37.0} \\
\midrule
Pre-edited && 0.50&0.50&15.2&0.40 & 0.50 & 14.3 &0.40&0.50&14.4& 26.5 & 26.7 & 28.0 &27.0 & 26.2 & 27.0& 27.8 & 26.5 & 26.0 \\
\midrule
FT& \multirow{10}{*}{\rotatebox{90}{GPT-J-6B}} 
& 45.0 & 5.00 & 4.00 & 0.00 & 0.00 & 0.00 & 0.00 & 0.00 & 0.00 
& 40.0 & 8.00 & 12.0 & 0.00 & 0.00 & 0.00 & 0.00 & 0.00 & 0.00 \\
RLEdit&  
& 80.0 & 36.0 & 5.50 & 0.00 & 0.00 & 0.00 & 0.00 & 0.00 & 0.00 
& 81.0 & 58.0 & 19.0 & 0.00 & 0.00 & 0.00 & 0.00 & 0.00 & 0.00 \\
RECT&   
& 82.0 & 52.0 & 9.50 & 0.00 & 0.00 & 0.00 & 0.00 & 0.00 & 0.00 
& 84.0 & 68.0 & 24.7 & 0.00 & 0.00 & 0.00 & 0.00 & 0.00 & 0.00 \\
PRUNE&  
& 87.0 & 54.0 & 10.6 & 0.00 & 0.00 & 0.00 & 0.00 & 0.00 & 0.00 
& 88.0 & 74.0 & 25.6 & 0.00 & 0.00 & 0.00 & 0.00 & 0.00 & 0.00 \\
AdaEdit&  
& 89.0 & 58.0 & 10.0 & 0.00 & 0.00 & 0.00 & 0.00 & 0.00 & 0.00 
& 86.0 & 71.0 & 24.8 & 0.00 & 0.00 & 0.00 & 0.00 & 0.00 & 0.00 \\
\cmidrule(lr){3-20} 
SimIE&  
& 96.4 & 72.1 & 12.8 & 0.00 & 0.00 & 0.00 & 0.00 & 0.00 & 0.00 
& 94.5 & 88.0 & 27.5 & 0.00 & 0.00 & 0.00 & 0.00 & 0.00 & 0.00 \\
EMMET&  
& 98.8 & 73.0 & 13.5 & 0.00 & 0.00 & 0.00 & 0.00 & 0.00 & 0.00 
& 98.0 & 92.0 & 28.8& 94.3 & 87.5 & 26.7& 87.5 & 82.6 & 22.8\\
PMET&   
& 98.5 & 72.5 & 13.5 & 85.0 & 50.0 & \textbf{7.80} & 30.0 & 18.0 & 2.50 
& \underline{98.0} & 93.0 & 29.4 & 89.0 & 81.0 & \textbf{28.0} & 0.00 & 0.00 & 0.00 \\
MEMIT&  
& 97.8 & 71.0 & \underline{13.8} & 74.5 & 40.4 & \underline{7.60} & 27.2 & 15.5 & 2.30 
& 96.5 & 92.0 & \textbf{30.1} & 95.0 & 89.0 & 24.0 & 0.00 & 0.00 & 0.00 \\
AlphaEdit& 
& \textbf{98.9} &\underline{73.5} & 13.7 & \underline{95.9} & \underline{50.9} & 6.90 & \underline{82.6} & \underline{40.3} & \underline{3.90} 
& \textbf{98.0} & \textbf{94.0} & 29.5 & \underline{96.5} & \underline{92.0} & 26.0 & \underline{88.3} & \underline{83.1} & \underline{22.8} \\
BetaEdit& 
& \underline{98.8} & \textbf{74.0} & \textbf{13.9} & \textbf{96.6} & \textbf{52.8} & 6.80 & \textbf{86.0} & \textbf{41.8} & \textbf{4.10} 
& 97.5 & \underline{93.0} & \underline{29.6}& \textbf{98.9} & \textbf{95.1} & \underline{26.7} & \textbf{92.5} & \textbf{86.5} & \textbf{24.7} \\
\midrule
Pre-edited &&0.40&0.20&22.3& 0.30 & 0.30 & 21.6&0.30&0.40&21.8& 36.5 & 36.9 & 37.5& 37.2 & 36.6 & 38.5& 37.8 & 37.4 & 37.2 \\
\midrule
FT& \multirow{10}{*}{\rotatebox{90}{LLaMA3-8B-Instruct}} 
& 48.0 & 5.90 & 8.70 & 0.00 & 0.00 & 0.00 & 0.00 & 0.00 & 0.00 
& 42.0 & 8.00 & 12.0 & 0.00 & 0.00 & 0.00 & 0.00 & 0.00 & 0.00 \\
RLEdit&  
& 80.0 & 38.0 & 14.7 & 0.00 & 0.00 & 0.00 & 0.00 & 0.00 & 0.00 
& 82.0 & 60.0 & 28.0 & 0.00 & 0.00 & 0.00 & 0.00 & 0.00 & 0.00 \\
RECT&   
& 84.0 & 58.0 & 17.0 & 0.00 & 0.00 & 0.00 & 0.00 & 0.00 & 0.00 
& 86.0 & 74.0 & 32.0 & 0.00 & 0.00 & 0.00 & 0.00 & 0.00 & 0.00 \\
PRUNE&  
& 88.0 & 60.0 & 17.5 & 0.00 & 0.00 & 0.00 & 0.00 & 0.00 & 0.00 
& 90.0 & 80.0 & 36.0 & 0.00 & 0.00 & 0.00 & 0.00 & 0.00 & 0.00 \\
AdaEdit&  
& 90.0 & 63.0 & 18.5 & 0.00 & 0.00 & 0.00 & 0.00 & 0.00 & 0.00 
& 88.0 & 76.0 & 32.0 & 0.00 & 0.00 & 0.00 & 0.00 & 0.00 & 0.00 \\
\cmidrule(lr){3-20} 
SimIE&  
& 96.0 & 68.0 & 20.7 & 0.00 & 0.00 & 0.00 & 0.00 & 0.00 & 0.00 
& 96.5 & 92.0 & 36.0 & 0.00 & 0.00 & 0.00 & 0.00 & 0.00 & 0.00 \\
EMMET&  
& 98.5 & 69.4 & 20.2 & 0.00 & 0.00 & 0.00 & 0.00 & 0.00 & 0.00 
& 98.5 & 94.0 & \textbf{40.5}  & 0.00 & 0.00 & 0.00  & 0.00 & 0.00 & 0.00 \\
PMET&   
& \underline{98.8} & 63.0 & \textbf{21.6} & 82.0 & 60.3 & 12.8 & 0.00 & 0.00 & 0.00 
& \textbf{98.8} & \underline{94.5} & 39.4 & 91.0 & 87.0 & 41.0 & 80.1 & 74.3 & 37.5 \\
MEMIT&  
& \textbf{99.3} & 61.0 & \underline{21.5} & \underline{87.1} & \textbf{65.4} & \underline{14.4} & 0.00 & 0.00 & 0.00 
& \underline{98.5} & 94.0 & 39.5 & \underline{97.0} & \underline{93.1} & \underline{43.5} & \underline{93.2} & \underline{89.3} & \textbf{44.2} \\
AlphaEdit& 
& 97.3 & \textbf{71.5} & 19.4  & 0.00 & 0.00 & 0.00 & 0.00 & 0.00 & 0.00 
& 86.0 & 81.0 & 40.1 & 84.2 & 79.0 & 39.3 & 0.00 & 0.00 & 0.00 \\
BetaEdit& 
& 96.8 & \underline{70.4} & 19.0 & \textbf{89.0} & 62.4 & \textbf{15.1} & \textbf{73.2} & \textbf{57.4} & \textbf{10.2} 
& 98.0 & \textbf{94.5} & \underline{40.3} &\textbf{ 97.4} &\textbf{ 93.2} & \textbf{44.2} & \textbf{96.6} & \textbf{92.4} & \underline{43.9} \\
\bottomrule
\end{tabular}%
\normalsize
\caption{
    Editing performance across 300, 5,000, and 10,000 sequential edits on \textsc{CounterFact} and \textsc{ZsRE}. 
    Higher values are better for all metrics (efficacy, generality and specificity). 
    Best and second-best results are bolded and underlined, respectively.
}
\label{tab:results}
\end{table*}

\paragraph{LLMs, Datasets, and Evaluation Metrics.}  
We evaluate our approach on three decoder-only language models: Qwen3-4B-Instruct-2507, GPT-J-6B, and LLaMA3-8B-Instruct. Following prior practice \cite{fangalphaedit}, we apply edits to layers \{4,5,6,7,8\} for LLaMA3 and Qwen3, and to layers \{3,4,5,6,7,8\} for GPT-J. 

We assess editing performance on two standard factual editing benchmarks: \textsc{CounterFact}~\cite{meng2022locating} and \textsc{ZsRE}~\cite{levy2017zero}. Performance is measured using three standard metrics:  
\textbf{Efficacy (Eff.)}, which measures whether the edited model assigns the highest probability to the target output \( y \) for the input \( x \) (i.e., \( y = \arg\max_{y'} P(y' \mid x) \));  
\textbf{Generality (Gen.)}, which evaluates whether the edited model generalizes to paraphrased inputs \( x_p \) (i.e., \( y = \arg\max_{y'} P(y' \mid x_p) \));  
and \textbf{Specificity (Spe.)},  which measures whether the edited model preserves unrelated facts. Formally, for an unrelated input   $  x_s  $   and its corresponding output   $  y_s  $  , it requires that   $  y_s = \arg\max_{y'} P(y' \mid x_s)  $.

\paragraph{Compared Baselines.}  
We compare against a range of state-of-the-art model editing methods, categorized as follows: (i) fine-tuning with LoRA (\textsc{FT})~\cite{hulora}; (ii) hypernetwork-based editing (\textsc{RLEdit})~\cite{lireinforced}; (iii) locate-then-edit approaches, including \textsc{MEMIT}~\cite{mengmass}, \textsc{PMET}~\cite{li2024pmet}, \textsc{EMMET}~\cite{gupta2024unified}, and \textsc{AlphaEdit}~\cite{fangalphaedit}; and (iv) post-hoc adjustment techniques for per-step weight update, including \textsc{RECT}~\cite{gu2024model}, \textsc{PRUNE}~\cite{maperturbation}, \textsc{AdaEdit}~\cite{li2025adaedit}, and \textsc{SimIE}~\cite{guotowards}. All locate-then-edit methods are adapted to the sequential editing setting using \textit{history-aware} regularization.

\begin{figure*}[!htbp]
  \centering
  \includegraphics[width=0.95\linewidth]{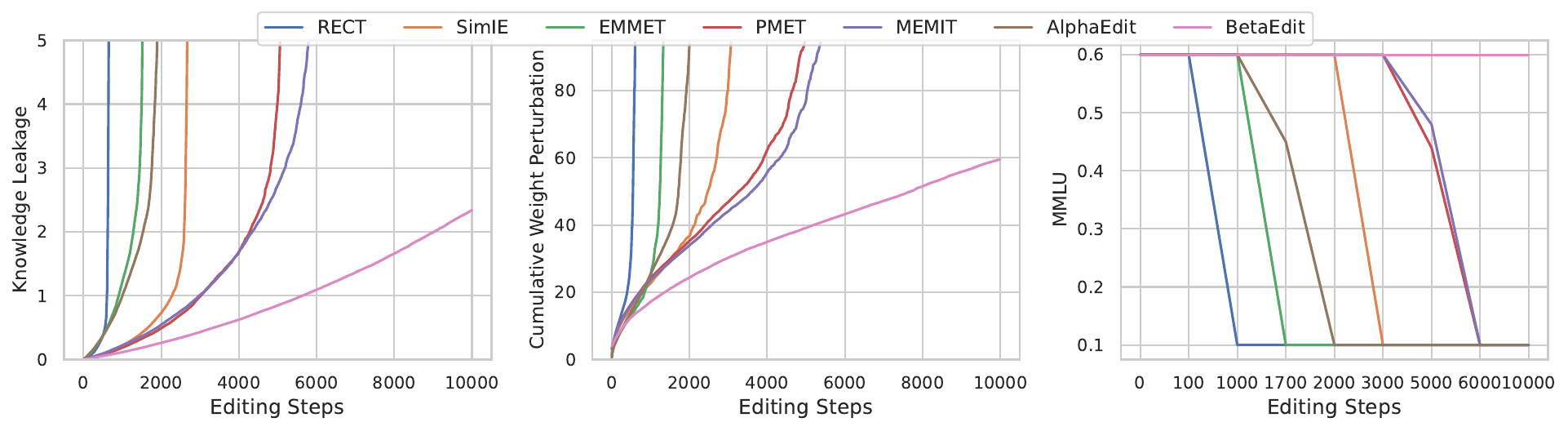}
  \caption{Evolution of model behavior during sequential editing on LLaMA3 using the CounterFact dataset: (left) knowledge leakage, (center) cumulative weight perturbation, and (right) MMLU performance, all measured as the number of edits increases from 0 to 10,000.}
  \label{fig:llama_cf_kl_weight_mmlu}
\end{figure*}

\subsection{Editing Performance}
We evaluate editing performance at three scales—300, 5,000, and 10,000 sequential edits—to more comprehensively assess the scaling behavior of existing editors. Unlike some baselines that use large batch sizes (e.g., 20 or 100 \cite{lireinforced,fangalphaedit}), we adopt the most challenging setting with a batch size of 1. Table~\ref{tab:results} summarizes editing performance across three models and two datasets. We group editors into two categories:  
(1) \textit{history-agnostic editors}, which perform each edit independently (e.g., FT, RECT and AdaEdit);  
and (2) \textit{history-aware editors}, which explicitly preserve the editing history during new updates (e.g., SimIE, EMMET and \textsc{BetaEdit}).
\begin{figure}[!htbp]
  \centering
  \includegraphics[width=1\linewidth]{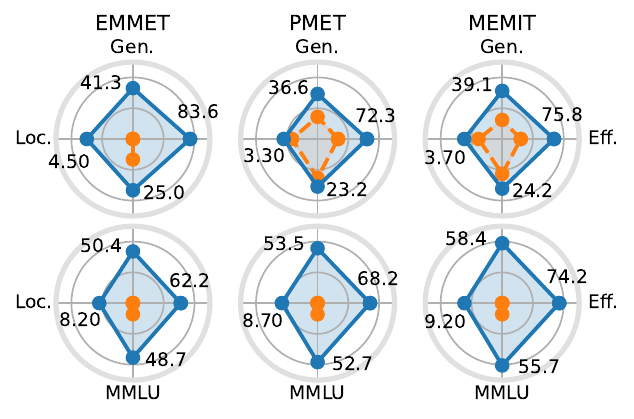}
  \caption{Editing performance and MMLU performance over 10,000 sequential edits with (blue) and without (orange) the null-space constraint on editing history, evaluated on GPT-J (top) and LLaMA3 (bottom) using the CounterFact dataset.}
  \label{fig:betaedit_ns}
\end{figure}

As shown in Table~\ref{tab:results}, history-aware editors consistently outperform the history-agnostic baselines. Notably, \textsc{BetaEdit} achieves the best overall performance, particularly under large-scale sequential editing. Most baselines degrade catastrophically beyond 5,000 edits, whereas \textsc{BetaEdit} remains effective even at 10,000 edits. We attribute this robustness to \textsc{BetaEdit}'s superior preservation of the model’s general capabilities. Once these language capabilities are compromised (i.e., models are corrupted), subsequent editing becomes ineffective or even infeasible~\cite{gupta2024model}. Figure~\ref{fig:llama_cf_kl_weight_mmlu} presents three key metrics as the number of edits increases from 0 to 10,000: (i) knowledge leakage, (ii) cumulative weight perturbation, and (iii) MMLU performance (Macro-F1 score). \textsc{BetaEdit} effectively suppresses knowledge leakage and limits cumulative parameter drift, thereby preserving the model’s general capabilities.  

Among the two recipes in \textsc{BetaEdit}, knowledge leakage control is specifically designed for null-space-based editing methods, while null-space-based history-aware regularization is a general-purpose technique that effectively minimizes interference with historical edits during new updates, thereby better preserving both editing performance and the model’s general capabilities—as established by our theoretical analysis in Section~\ref{sec:analysis}. To further evaluate its effectiveness, we apply this technique to the baseline editors EMMET, PMET, and MEMIT. As shown in Figure~\ref{fig:betaedit_ns}, when enhanced with null-space-based history-aware regularization, all editing methods remain effective under 10,000 sequential edits without corrupting the model.






\begin{figure}[!htbp]
  \centering
  \includegraphics[width=0.95\linewidth]{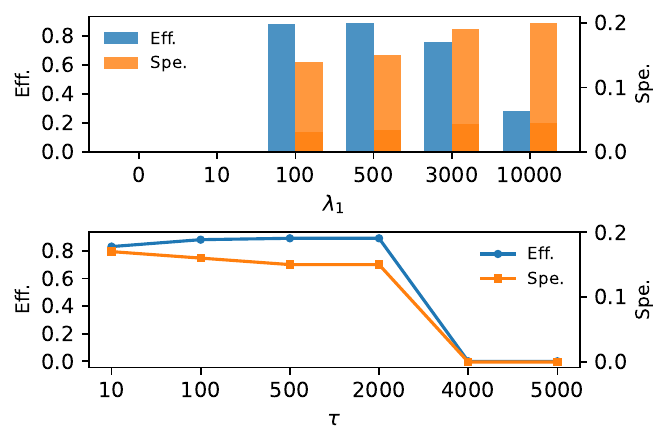}
\caption{(Top) Impact of the penalty coefficient $\lambda_1$ for knowledge leakage; (Bottom) Impact of the refresh period $\tau$ for the null-space projector $\mathbf{P}_t$. Performance is measured in terms of editing efficacy (Eff., left y-axis) and specificity (Spe., right y-axis) on \textsc{LLaMA}3 over 5,000 sequential edits using the CounterFact dataset.}
  \label{fig:betaedit_hp}
\end{figure}

\subsection{Ablation Study and Hyperparameter Analysis}
\paragraph{Ablation Study.} We evaluate \textsc{BetaEdit} under two ablated settings: (1) without the knowledge leakage (KL) penalty term, and (2) without the null-space (NS) constraint for preserving prior edits (denoted as \textit{w/o KL} and \textit{w/o NS}, respectively). The ablation results are shown in Table \ref{tab:ablation} (top: 5k edits; bottom: 10k edits). Two key observations emerge:

\begin{table*}[htbp]
\centering
\small
\setlength{\tabcolsep}{4pt}
\begin{tabular}{l l c c c c c cc c c c c cc}
\toprule
\multirow{3}{*}{\textbf{Method}} & 
& \multicolumn{6}{c}{\textbf{CounterFact}} 
& \multicolumn{6}{c}{\textbf{ZsRE}} \\
\cmidrule(lr){3-8} \cmidrule(lr){9-14}
&&\multicolumn{2}{c}{\textbf{Qwen3}}&\multicolumn{2}{c}{\textbf{GPT-J}}&\multicolumn{2}{c}{\textbf{LLaMA3}}&\multicolumn{2}{c}{\textbf{Qwen3}}&\multicolumn{2}{c}{\textbf{GPT-J}}&\multicolumn{2}{c}{\textbf{LLaMA3}}\\
\cmidrule(lr){3-4} \cmidrule(lr){5-6}  \cmidrule(lr){7-8} \cmidrule(lr){9-10} \cmidrule(lr){11-12} \cmidrule(lr){13-14} 
& 
& \textbf{Eff.} & \textbf{MMLU}& \textbf{Eff.} & \textbf{MMLU}& \textbf{Eff.} & \textbf{MMLU}& \textbf{Eff.} & \textbf{MMLU}& \textbf{Eff.} & \textbf{MMLU}& \textbf{Eff.} & \textbf{MMLU} \\
\midrule
\textit{w/o KL}& \multirow{3}{*}{\rotatebox{90}{5k}}&0.00&10.0&95.9&24.2&0.00&10.0&0.00&10.0&96.4&24.5&0.00&10.0 \\
\textit{w/o NS}& &0.00&10.0&96.2&25.6&0.00&10.0&92.7&59.3&96.4&25.2&0.00&10.0\\
\textit{w/ Both}&  &90.3&60.9&96.6&25.0&89.0&59.0&96.7&61.8&98.9&25.0&97.4&60.5 \\
\midrule
\textit{w/o KL}& \multirow{3}{*}{\rotatebox{90}{10k}}&0.00&10.0 &82.2&21.5&0.00&10.0&0.00&10.0&90.1&23.0&0.00&10.0\\
\textit{w/o NS}& &0.00&10.0&81.4&22.3&0.00&10.0&0.00&10.0&90.4&23.2&0.00&10.0\\
\textit{w/ Both}&  &78.5&59.2&86.0&23.6&73.2&57.4&92.1&59.6&92.5&23.8&96.6&59.1 \\
\bottomrule
\end{tabular}%
\normalsize
\caption{
    Ablation study of \textsc{BetaEdit}, evaluating the impact of removing the knowledge leakage (KL) penalty term or the null-space (NS) constraint for history edits. 
    Results are shown for 5,000 edits (top block) and 10,000 edits (bottom block), reporting editing efficacy (\%) and  Macro-F1 score (\%) of the MMLU task. 
}
\label{tab:ablation}
\end{table*}

First, penalizing knowledge leakage is essential across both editing scales (5k and 10k), serving as a fundamental mechanism to protect pre-trained knowledge. Second, the null-space constraint on historical edits significantly enhances editing efficacy and preserves the model’s general capabilities—particularly at 10k edits, where KL-only variants (\textit{w/o NS}) typically achieve near-zero editing efficacy and corrupt the model’s general capabilities. This underscores the critical role of protecting prior edits: it not only mitigates interference with editing history but also substantially curbs cumulative parameter perturbation—a key factor, as shown in Section \ref{sec:analysis}, for preserving the model’s general functionality during long sequences of edits.
\paragraph{Hyperparameter Analysis.}
We study two key hyperparameters in \textsc{BetaEdit}: the knowledge leakage penalty coefficient $\lambda_1$ and the null-space projector refresh period $\tau$.

The top panel of Figure \ref{fig:betaedit_hp} shows the effect of varying $\lambda_1 \in \{0, 10, 100, 500, 3000, 10000\}$ on \textsc{LLaMA}3 under 5,000 sequential CounterFact edits. Small $\lambda_1$ (e.g., 0 or 10) leads to severe knowledge leakage, collapsing both efficacy and specificity near zero. Larger values better preserve pretrained knowledge and improve both metrics, but excessively large $\lambda_1$ (e.g., 10,000) over-constrains updates, limiting editability and degrading efficacy despite high specificity—highlighting a trade-off between editability and knowledge retention.

The bottom panel of Figure \ref{fig:betaedit_hp} examines the refresh interval $\tau$, i.e., how often the null-space projector $\mathbf{P}_t$ is recomputed. With $\tau = 5000$ (no refresh), \textsc{BetaEdit} relies only on standard history-aware regularization (\textit{w/o NS}) and fails to sustain performance: both metrics drop near zero. Since recomputing $\mathbf{P}_t$ is costly, a moderate $\tau$ (e.g., 1,000) suffices to maintain strong editing performance while avoiding unnecessary computational overhead.


\section{Limitations}
\label{sec:limit}
Existing editing methods typically perform edits at the \emph{subject} token, which aligns well with standard editing benchmarks such as CounterFact and ZsRE—datasets where each edit involves a distinct subject. In this setting, edits are generally non-conflicting, and our theoretical analysis of \textit{history-aware} regularization holds reasonably well. However, the non-conflicting assumption can be significantly violated in two practical scenarios: (1) when edits are not anchored at the subject token, or (2) when multiple edits share the same subject. As shown in Table \ref{tab:tok}, anchoring edits at the last token drastically reduces the effectiveness of history-aware regularization. On LLaMA3, edit efficacy already drops to 54.7\% after just 300 sequential edits, and by 2,000 edits the model is effectively corrupted—far below the 10,000-edit capacity achievable with subject-token anchoring. This is because the last token (e.g., ``is'') frequently appears across many edit prompts; anchoring knowledge update at this position introduces substantial interference among edits.

This reveals a key limitation of our approach: its dependence on subject-token anchoring imposes practical constraints. Specifically, it struggles in settings involving (i) multiple edits targeting the same subject \cite{dong2025memit}, or (ii) edits with complex or long-form prompts that deviate from simple subject–relation–object triplets and may involve multiple subjects \cite{jianganyedit,zeng2025docmedit}.


\begin{table}[htbp]
\centering
\footnotesize 
\setlength{\tabcolsep}{4pt}
\begin{tabular}{l l c c c c c cc}
\toprule
\multirow{2}{*}{\textbf{Method}}
&&\multicolumn{2}{c}{\textbf{Qwen3}}&\multicolumn{2}{c}{\textbf{GPT-J}}&\multicolumn{2}{c}{\textbf{LLaMA3}}\\
\cmidrule(lr){3-4} \cmidrule(lr){5-6}  \cmidrule(lr){7-8}
& 
& \textbf{Eff.} & \textbf{MMLU}& \textbf{Eff.} & \textbf{MMLU}& \textbf{Eff.} & \textbf{MMLU} \\
\midrule
\textit{Last}& \multirow{4}{*}{\rotatebox{90}{300}}&22.4&47.3&38.7&18.3&26.3&46.7\\
\textit{Last+H}& &50.8&52.0&65.4&20.3&54.7&49.5\\
\textit{Sub.}&  &82.4&60.2&96.0&25.1&79.7&57.0 \\
\textit{Sub.+H}&  &98.0&62.9&98.8&25.3&96.8&60.1 \\
\midrule
\textit{Last}& \multirow{4}{*}{\rotatebox{90}{2000}}&0.00&10.0&0.00&10.0&0.00&10.0\\
\textit{Last+H}& &0.00&10.0&0.00&10.0&0.00&10.0\\
\textit{Sub.}& &0.00&10.0&0.00&10.0&0.00&10.0\\
\textit{Sub.+H}&  &97.1&62.4&98.0&25.0&95.6&59.8 \\
\bottomrule
\end{tabular}%
\normalsize
\caption{
    Effectiveness of history-aware regularization (\textit{H}) under different anchoring strategies: last token (\textit{Last}) vs. subject token (\textit{Sub.}). Results are reported for 300 (top) and 2,000 (bottom) sequential edits on the CounterFact dataset.
}
\label{tab:tok}
\end{table}
\section{Conclusion}
\label{sec:conclusion}

We have uncovered a fundamental gap between the theoretical promise and practical behavior of null-space-based model editing: despite being designed to preserve pre-trained knowledge perfectly, these methods suffer from \textit{knowledge leakage} due to the use of approximate null spaces, leading to catastrophic failure in long-horizon sequential editing.  Beyond identifying this issue, we provide a principled theoretical analysis showing that \textit{history-aware} regularization serves a dual role—it not only protects previously edited facts but also implicitly constrains overall parameter drift, thereby preserving the model’s general capabilities.  

Building on these insights, we propose \textsc{BetaEdit}, a refined editing framework that explicitly mitigates knowledge leakage by reintroducing a dedicated penalty term, while seamlessly integrating history-aware updates into the null-space paradigm. Experiments across multiple large language models and standard benchmarks show that \textsc{BetaEdit} maintains high editing performance and effectively preserves the model’s general capabilities—even under up to 10,000 sequential edits, a regime in which prior methods typically collapse—demonstrating its effectiveness for long-horizon model editing.

\bibliographystyle{named}
\bibliography{ijcai26}

\end{document}